\pdfoutput=1
\documentclass[11pt]{article}

\usepackage[]{acl}

\usepackage{times}
\usepackage{latexsym}

\usepackage{cleveref}

\usepackage[T1]{fontenc}
\usepackage[utf8]{inputenc}

\usepackage{microtype}

\usepackage{inconsolata}

\usepackage{graphicx}
\usepackage{comment}

\usepackage{pifont}% http://ctan.org/pkg/pifont
\newcommand{\cmark}{\ding{51}}%
\newcommand{\xmark}{\ding{55}}%

\newcommand{\camred}[1]{{#1}} % Remove comment to display all new text in camera-ready version as normal text

\title{Do Pre-Trained Language Models Detect and Understand Semantic Underspecification? \emph{Ask the DUST!}}

\author{Frank Wildenburg \\
  College of Informatics \\
  University of Amsterdam \\
  \texttt{f.c.l.wildenburg@uva.nl} \\\And
  Michael Hanna \\
  ILLC \\
  University of Amsterdam \\
  \texttt{m.w.hanna@uva.nl} \\\And
  Sandro Pezzelle \\
  ILLC \\
  University of Amsterdam \\
  \texttt{s.pezzelle@uva.nl} \\\\}

\begin{document}
    \maketitle
\begin{abstract}

In everyday language use, speakers frequently utter and interpret sentences that are \emph{semantically underspecified}, namely, whose content is insufficient to fully convey their message or interpret them univocally. For example, to interpret the underspecified sentence ``Don't spend too much'', which leaves implicit what (not) to spend, additional linguistic context or outside knowledge is needed. In this work, we propose a novel \textbf{D}ataset of semantically \textbf{U}nderspecified \textbf{S}entences grouped by \textbf{T}ype (\textbf{DUST}) and use it to study whether pre-trained language models (LMs) correctly identify and interpret underspecified sentences. We find that newer LMs are reasonably able to identify underspecified sentences when explicitly prompted. However, interpreting them correctly is much harder for any LMs. Our experiments show that when interpreting underspecified sentences, LMs exhibit little uncertainty, contrary to what theoretical accounts of underspecification would predict. Overall, our study reveals limitations in current models'  processing of sentence semantics and highlights the importance of using naturalistic data and communicative scenarios when evaluating LMs' language capabilities.
\end{abstract}

\section{Introduction}
Speakers can almost effortlessly deal with
\emph{semantic underspecification}, a widespread phenomenon that
occurs when a linguistic signal does not fully convey all the information required for communication to succeed~\cite{frisson_semantic_2009, harris_what_2020}.
This is possible because, in a normal state of affairs, humans have access to further linguistic or extra-linguistic information coming from the conversation, the surrounding context, or shared knowledge. If this is the case, speakers will have no trouble understanding the sentences ``I'll meet you there'' or ``I saw you on the hill with the telescope'', even though these sentences leave underspecified where ``there'' is, or which interlocutor had ``a telescope''.
% when ``tomorrow'' is, and 
% who or  what has ``three legs''
% For example, most speakers will have no trouble understanding the sentences ``We'll meet here tomorrow'' or ``I left the chair with three legs'', even though these sentences leave underspecified where ``here'' is, or 
% when ``tomorrow'' is, and 
% who or  what has ``three legs''. 
It has been argued that underspecification and the related phenomenon of ambiguity are not a hindrance in language, but rather a desirable feature of human language communication~\cite{piantadosi_communicative_2012};
% For example, 
they allow for more concise utterances, which makes language more efficient.
% they make language more efficient, by making utterances more concise. 
Indeed, humans excel at making inferences \citep{grice1969utterers}, which is less cognitively taxing compared to 
% producing sounds
articulating speech~\cite{levinson_presumptive_2000}.

\begin{figure}[t!]
    \centering
    \resizebox{\columnwidth}{!}{%
    \begin{tabular}{l l l}

        \hline
         \textbf{S1} & Don't spend too much. & \\
         \textbf{S2} & Don't spend too much cash. & \\
         & & \\
         
         Exp. 1 & \textit{Is \textbf{S2} more specified than \textbf{S1}?} & \textcolor{green}{\cmark}\\ \hline 
         %& & 
         %\\
         \textbf{S1} & The bag is on the chair. It is green & \\
         \textbf{S2} & The bag is on the chair. The chair & \\
         & is green. & \\
         & \\
         
         Exp. 2 & \textit{Does \textbf{S1} mean the chair is green?} & \textcolor{blue}{?}\\
         & \textit{Does \textbf{S2} mean the chair is green?} & \textcolor{green}{\cmark}\\
         & \textit{Does \textbf{S1} mean the chair isn't green?} & \textcolor{blue}{?} \\
         & \textit{Does \textbf{S2} mean the chair isn't green?} & \textcolor{red}{\xmark}\\ \hline
         
    \end{tabular}
    }
    \caption{
    % Underspecified sentences (\textbf{S1}) may license more interpretations than more specified counterparts (\textbf{S2}). 
    In experiment 1 (top), we test if LMs % recognize
    distinguish underspecified sentences (\textbf{S1}) from more specified counterparts (\textbf{S2})
    % underspecification 
    when explicitly prompted. In experiment 2 (bottom), we test 
    % how LMs interpret 
    if LMs correctly interpret \textbf{S1} and \textbf{S2} in a more naturalistic communicative setting.}
    % underspecified sentences.}
    \label{fig:illustration}
\end{figure}

While humans are good at dealing with semantically underspecified language by leveraging additional information, how pre-trained transformer language models (LMs) behave when faced with this phenomenon is an open question. Despite the growing literature exploring the semantic capabilities of last-generation LMs~\cite{ettinger_what_2020,rogers-etal-2020-primer,hanna-etal-2023-language}, little attention has been paid to this problem. Furthermore, 
the few previous works
% previous work 
generally did not distinguish between the various facets of ambiguity and underspecification~\cite{liu_were_2023} or only focused on ambiguity~\cite{stengeleskin_semanticparsing_2024, ortega-martin_linguistic_2023, fantechi_rule-based_2023}.

% This is although handling underspecification is an important ability for LMs. 
However, handling semantically underspecified language is critical for LMs. Since underspecified sentences can license multiple interpretations, choosing one arbitrarily can lead to undesired or even harmful consequences for communication~\cite[see ][for an in-depth discussion of this issue]{hutchinson_underspecification_2022,pezzelle_dealing_2023}.  %\camred{For example, when translating a sentence with gender-neutral pronouns (i.e. a sentence in which gender is left underspecified) into a language without such pronouns, NLP systems should default towards a (potentially stereotypical), but rather, interpret the underspecification or ask for clarification. \cite{farkas2022bias}.}
\camred{Correctly processing underspecified language could have real-world consequences for NLP systems. For example, if an embodied virtual assistant
% such a system 
were to misinterpret a question like ``Can I hang the painting with the cup?'', there could be risks: while the answer may be \textit{yes} if the cup is the subject of the painting, the system should answer \textit{no} if this is not the case. In machine translation, models should carefully handle underspecification; for example, when translating ``they'', models must determine whether the pronoun refers to a group of people or an individual of unknown, nonbinary, or intentionally underspecified gender.
}
Therefore, LMs should (1) recognize semantically underspecified inputs and (2) interpret them appropriately, ideally 
with no biases toward a default reading.

% without choosing a biased or default reading.
% potentially leading to harmful misunderstandings between communicators -- e.g. in the case of \Cref{fig:illustration}, the underspecified sentence may lead to the wrong bag being picked up. 
% For such reasons, LMs need to be able to deal with semantic underspecification to appropriately perform language tasks~\citep{pezzelle_dealing_2023}. %hutchinson_underspecification_2022

%In this work, 
We build on and extend previous work by asking two new questions: (1) to what extent 
can LMs detect if a sentence is (under)specified? (2) How do LMs interpret such underspecified sentences compared to more specified counterparts? To this end, we introduce \textbf{DUST}\footnote{We release DUST and code for our experiments here: \href{https://github.com/frank-wildenburg/DUST}{https://github.com/frank-wildenburg/DUST}}, a \textbf{D}ataset of semantically \textbf{U}nderspecified \textbf{S}entences (and more specified counterparts) grouped by the \textbf{T}ype of underspecification they belong to, and propose a suite of experiments 
to answer the questions above
using DUST. To categorize
%distinguish between different types 
instances of underspecification, we build on
% formal categorization 
\citeposs{egg_semantic_2010} proposed taxonomy.

Our experiments and analysis show that (1) distinguishing semantically underspecified sentences from specified counterparts is not a trivial task for current LMs. While newer, better-performing models achieve reasonable performance when given explicit instructions, other models fare only slightly better than random.
% to a reasonable when explicitly prompted
% only larger, best-performing LMs can distinguish underspecified sentences from specified counterparts to a reasonable when explicitly prompted }
% find 
% that, while better-performing models can distinguish underspecified sentences from specified counterparts reasonably well when explicitly prompted, their ability to do so is smaller than in comparable experiments that use more commonly tested language features such as sentiment. 
Moreover, (2) when asked to interpret underspecified sentences in a more naturalistic communicative scenario without explicit guidance, all LMs fall into the trap of interpreting them similarly to their more specified versions. This suggests that, in the absence of a specific prompt, these models assign biased or default interpretations to underspecified and ambiguous sentences.
% performance drops to a near-chance level for all LMs. 

%  \footnote{We will release our code and data.} code upon acceptance at: \url{github.com/anonymous/repo}

% Model performance also differs between types of underspecification and is correlated with the concreteness of constituent words.

Our findings confirm that current LMs, including the best-performing Llama 2~\cite{touvron_llama_2023} and Mistral~\cite{jiang_mistral_2023}, struggle with underspecified and ambiguous language~\cite[in line with][]{liu_were_2023}. This reveals more general limitations with the processing of sentence semantics. Moreover, our study highlights the importance of methodological choices, such as experimental setting, or the level of informativeness of prompts, in fine-grained evaluations of LMs' capabilities.

% Our findings indicate that, in line with the related phenomenon of ambiguity~\cite{liu_were_2023}, LMs find underspecification extremely challenging. Furthermore, care should be taken to represent the different facets of ambiguity and underspecification when evaluating the performance of LMs.

\section{Related work}
\label{sec:relatedwork}
\subsection{Semantic Underspecification} Semantic underspecification is a phenomenon in which the semantic material of a sentence ``leaves open'' possibilities for readers of the text that may then be ``filled in'' through non-linguistic information~\cite{zwicky_ambiguity_1975, frisson_semantic_2009, egg_semantic_2010, harris_what_2020}. The phenomenon has traditionally been studied through the lens of (formal) linguistics and semantics (\citealp{zwicky_ambiguity_1975, lappin_intensional_2000, egg_semantic_2010, harris_what_2020}, \textit{inter alia}), although it has also been studied in other fields, such as the neuroscience of language processing~\cite{frisson_semantic_2009} and information theory~\cite{piantadosi_communicative_2012, franzon_entropy_2022}.

%Earlier work on semantic underspecification considered the phenomenon from the perspective of (formal) semantics. For example, \citet{zwicky_ambiguity_1975} considered which tests linguistics can use to distinguish between underspecification and ambiguity, and in \citet{lappin_intensional_2000}, who gives an intensional parametric semantics for vague quantifiers through the use of an underspecified parametric interpretation. One particularly detailed investigation into underspecification is that of \citet{egg_semantic_2010}, which we describe in more detail in \Cref{sec:framework}.

Semantic underspecification is often studied in tandem or in contrast with ambiguity, a related phenomenon. The difference between the two is that underspecified sentences sometimes have only one reading (which may be clarified by non-linguistic information), whereas ambiguous sentences have multiple readings~\cite{zwicky_ambiguity_1975}. However, all ambiguous sentences are also underspecified; after all, some non-linguistic information will disambiguate between readings. Hence, underspecification can be seen as a generalization of ambiguity. Despite this, the terms are sometimes used interchangeably or in tandem; 
% M: not adding the page because it's just an online stanford encyclopedia of philosophy page (the quote is in the intro)
\citet{sennet_ambiguity_2023} points out that ``often simple underspecificity will suffice for a charge of ambiguity''.

\citet{egg_semantic_2010} gives a detailed categorization of underspecification, grouping instances into four types based on whether the instance's constituent parts comprise the same semantic value across its readings, and whether it is possible to give a single syntactic analysis for all the readings. \camred{As this classification allows us to
explore the interaction
% distinguish 
between the semantic and syntactic dimensions of underspecification and their
effects
% influence 
on LMs, we use it as the theoretical backbone to build our dataset.}
% basis of our dataset.}

\subsection{Semantic Underspecification in NLP}
NLP has long studied problems around semantic underspecification. Early work explicitly modeled underspecification by creating formal, symbolic representations of underspecified sentences that captured each sentence's potential meanings, without generating them (\citealp{poesio_ambiguity_1994,niehren_uniform_1997,pinkal_semantic_1999}, \textit{inter alia}). NLP systems could then use such representations to make processing sentences more tractable, despite their potentially numerous interpretations~\cite{wahlster_mobile_2000}.

% More recent work 
Another line of work
focuses instead 
on identifying or resolving underspecification.~\citet{stengeleskin_semanticparsing_2024}, for example, identify the meanings of ambiguous sentences by training a model to map from such sentences to formal representations of their multiple potential meanings.~\citet{berzak_you_2015} 
% instead 
train a model to resolve underspecification and determine the correct reading of an ambiguous caption, given an accompanying image. More generally, many classic NLP tasks, such as word sense disambiguation and coreference resolution, involve resolving a word's meaning in a context where it is underspecified. Where underspecification cannot be resolved, studies have tried to identify or generate % plausible 
clarifications or clarification questions~\citep{roth_semeval-2022_2022,testoni2024asking}.

% Another line of research seeks to understand the underspecification processing abilities of pre-trained language models. Such models generally do not explicitly model underspecification, and receive no explicit signal regarding underspecification. It is thus unclear how such models process underspecification, and if they process it correctly. The fact that ambiguous or underspecified instances are systematically excluded in the curation of benchmarks~\citep{liu_were_2023, stengeleskin_semanticparsing_2024} makes answering this question challenging.

Some recent work has addressed ambiguity and underspecification in the context of pre-trained LMs. Considering multi-modal models,~\citet{prasad2023rephrase} find that vision-language architectures often struggle with underspecified inputs; specifying inputs improves performance.~\citet{pezzelle_dealing_2023} reports similarly negative results, discovering that CLIP~\cite{radford2021learning} sometimes prefers invalid but highly specified captions to valid but underspecified ones. Multi-modal models' challenges with underspecification concern not only performance but also ethics:~\citet{hutchinson_underspecification_2022} warn that image generation models might rely on social biases to fill in underspecified details.% in their inputs.

% In a uni-modal, text-only context, most previous research focuses on ambiguity rather than underspecification more broadly. For example,~\citet{ortega-martin_linguistic_2023} and \citet{fantechi_rule-based_2023} study Chat-GPT's ability to explicitly identify ambiguity, and report mixed results. Most pertinently,~\citet{liu_were_2023} study how pre-trained LMs process ambiguous sentences, and find that they are unable to use context to infer which potential reading of an ambiguous sentence is correct. 

% Current evidence suggests that underspecification is challenging for pre-trained models. Considering multi-modal models,~\citet{prasad2023rephrase} find that vision-language models often struggle with underspecified inputs; specifying inputs yields improved performance.~\citet{pezzelle_dealing_2023} reports similarly negative results, discovering that CLIP~\cite{radford2021learning} sometimes prefers invalid but highly specified captions to valid but underspecified ones. Multi-modal models' challenges with underspecification concern not only performance but also ethics:~\citet{hutchinson_underspecification_2022} warn that image generation models might rely on social biases to fill in underspecified details in their inputs.

Related questions have been studied in a uni-modal, text-only context as well, though existing work focuses on ambiguity, rather than underspecification more broadly. For example,~\citet{ortega-martin_linguistic_2023} and \citet{fantechi_rule-based_2023} study Chat-GPT's ability to explicitly identify ambiguity, and report mixed results. Most pertinently,~\citet{liu_were_2023} study how pre-trained LMs process ambiguous sentences, and find that they are unable to use context to infer which potential reading of an ambiguous sentence is correct. In the present work, we aim to expand the existing literature to cover not just ambiguity, but all types of underspecification.

\begin{table}
    \centering
    \begin{tabular}{r | l | l | l}
        \textbf{T} & \textbf{Phenomenon} & \textbf{\#S} & \textbf{Source}\\
        \hline
        1 & Logical Form & 35 & \hyperlink{cite.berzak_you_2015}{LAVA}\\
        & Ellipsis & 18 & \hyperlink{cite.berzak_you_2015}{LAVA}\\
        \hline
        2 & PP attachment amb. & 48 & \hyperlink{cite.berzak_you_2015}{LAVA}\\
        & VP attachment amb. & 60 & \hyperlink{cite.berzak_you_2015}{LAVA}\\
        & Conjunction amb. & 40 & \hyperlink{cite.berzak_you_2015}{LAVA}\\
        \hline
        3 & Referential amb. & 36 & \hyperlink{cite.berzak_you_2015}{LAVA}\\
        & Referential amb. & 273 & \hyperlink{cite.levesque_winograd_2012}{WSC}\\
        & \emph{Added compound} & 774 & \hyperlink{cite.roth_semeval-2022_2022}{CLAIRE}\\
        & \emph{Fused head} & 532 & \hyperlink{cite.roth_semeval-2022_2022}{CLAIRE}\\
        & \emph{Implicit reference} & 216 & \hyperlink{cite.roth_semeval-2022_2022}{CLAIRE}\\
        & \emph{Metonymic reference} & 91 & \hyperlink{cite.roth_semeval-2022_2022}{CLAIRE}\\
        %\hline
        %4 & Homonymy & 980 & \hyperlink{cite.wikisent}{WikiSent}\\
    \end{tabular}
    \caption{Number of sentences (\#S) and source per type (T) and phenomenon of underspecification in DUST. Phenomena containing underspecification without ambiguity are \textit{italicized}. `amb.' stands for ambiguity.}
    \label{tab:dataset_statistics}
\end{table}

\section{Dataset}

To study how LMs deal with semantically underspecified language, and since there currently exists no comprehensive resource on underspecification, we construct DUST, a \textbf{D}ataset of \textbf{U}nderspecified \textbf{S}entences by \textbf{T}ype, consisting of 2,123 %2,246 %3226 
English underspecified sentences and equally many specified counterparts, based on \citeposs{egg_semantic_2010} categorization; see \Cref{tab:dataset_statistics} for an overview. Below, we discuss the construction of the dataset by type; note that although \citeauthor{egg_semantic_2010}'s taxonomy includes 4 types of underspecification, we only include 3 in our dataset due to the features of one type (more details below). 
% we include no examples of the last type. 

\paragraph{Type 1}
\citet{egg_semantic_2010}'s first type consists of \textit{semantically and syntactically homogeneous expressions}, i.e., sentences with multiple readings that all share the same syntactic structure, and word/token-level meaning.
To cover this type of underspecification, we collect 53 sentences from the Language and Vision Ambiguities (LAVA) dataset~\citep{berzak_you_2015}, a multimodal dataset containing ambiguous sentences and visual data that disambiguated them. For each sentence in this dataset, we created a more specified counterpart using its visual disambiguations. An example of such a sentence is
\begin{quote}
    Andrei approached Danny; Yevgeni, too.
\end{quote}
which leaves underspecified whether Yevgeni approached Danny, or was approached by Andrei. A potential specified version of this sentence is
\begin{quote}
    Andrei approached Danny; Andrei approached Yevgeni, too.
\end{quote}

\paragraph{Type 2}
The second type consists of \textit{semantically but not syntactically homogeneous expressions}. The multiple readings of such expressions stem from the multiple ways of analyzing the expression's structure; different analyses can lead to different meanings. The LAVA dataset~\cite{berzak_you_2015} provides 108 sentences containing VP and PP attachment ambiguity and conjunction ambiguity, for example
\begin{quote}
    Andrei looked at the green bag with a telescope.
\end{quote}
which leaves ambiguous whether `with a telescope' attaches to `the green bag' or to `looked at'. It is thus unclear whether the bag contains a telescope, or was looked at using one. A more specified counterpart of this sentence might be
\begin{quote}
    Andrei looked at the green bag through a telescope.
\end{quote}

\paragraph{Type 3}

The third type consists of \textit{syntactically but not semantically homogeneous expressions}, which share a single syntactic structure but do not share the same semantic material in its constituent parts. It is unique in that, besides instances of referential ambiguity, all examples of type 3 in DUST are underspecified but not ambiguous. It is thus particularly interesting for our work, as underspecified but not ambiguous expressions are important for LMs to handle, but currently understudied.
Examples include deictic expressions and expressions that are underspecified due to missing information.

As examples of this type of underspecification, we first collect 89 sentences from LAVA that contain referential ambiguity or missing information. We then extend our collection with sentences from the CLArifying Insertions from REvision Edits (CLAIRE) dataset~\cite{roth_semeval-2022_2022}, which consists of wikiHow instructional texts, and revisions that clarify the original sentences by inserting additional information. We treat the pre-edit text as underspecified due to missing information, and the post-edit text as more specified. Due to the original authors' pre-processing, some pre-edit sentences are ungrammatical; we thus score pre-edit sentences' grammaticality with GRUEN \citep{zhu_gruen_2020} and exclude low-scoring sentences.

We also include the original 273 sentences from the Winograd Schema Challenge~\cite[WSC;][]{levesque_winograd_2012} in our dataset as examples of referential ambiguity. For each sentence from this dataset, we crafted a more specified counterpart by changing the gender or plurality of one of the antecedents in the sentence, removing the referential ambiguity. An example sentence of this type is
\begin{quote}
    Don't spend too much.
\end{quote}
which does not specify what should not be spent. A more specified counterpart might be
\begin{quote}
    Don't spend too much cash.
\end{quote}

\paragraph{Type 4}
\citeauthor{egg_semantic_2010}'s fourth and final type, consisting of expressions that are \textit{neither syntactically nor semantically homogeneous}, concerns phenomena that occur at the word level, such as homonymic expressions. 
While
% Hence, while 
a word (e.g. ``plant'') can have multiple syntactically and semantically distinct readings, most sentences that contain homonyms have readings that rely on the same syntactic structure. For example, ``he walked to the bank'' contains the homonym ``bank'', which could refer to a riverbank of a financial institution, but the syntactic structure of the sentence is the same across both readings. 

\camred{
% While 
Unlike the three types described above, the phenomena belonging to this type cannot be easily studied using an
% are an important source of underspecification, they  are not compatible with an
experimental setup based on minimal pairs---the one used in this work---where the two interpretations are embedded in two phrases or sentences that only differ by a single, minimal intervention (see~\Cref{sec:type4} for a preliminary exploration of the problem). For this reason, we do not include type 4 in our benchmark and leave a comprehensive exploration of it to future work.}

\section{Models}
%\camred{TODO: mention pure/fine-tuned}
We focus on pre-trained autoregressive models, including both 
older and newer (generally stronger) LMs
% stronger  (newer) and weaker models 
to consider the influence of general performance improvements on LMs' underspecification processing. As our experiments require sentence probabilities, we use only openly available models, which provide these. Specifically, we consider the following models, accessed using HuggingFace's Transformers library \cite{wolf2020huggingfaces}:\footnote{See \Cref{sec:model_exp_details} for more model details.}

\begin{itemize}
    \item GPT-2 XL \cite{radford_language_2019}, a 1.5 billion parameter decoder-only LM trained on a 40GB dataset of webpages.

    \item FLAN-T5 XXL \cite{chung_scaling_2022}, a 11 billion parameter encoder-decoder LM. FLAN-T5 is an enhanced version of T5, finetuned in a mixture of language modeling tasks.

    \item OPT-13b \cite{zhang_opt_2022}, a 13 billion parameter decoder-only transformer model.% trained to rougly match the performance and sized of the GPT-3 class of models.

    \item Llama 2 7b and 13b \cite{touvron_llama_2023}, 7 and 13 billion parameter\footnote{Due to compute limits, we only use the 13 billion parameter variant of Llama 2 in our second experiment} models trained on publicly available online data.

    \item Mistral 7b v0.1 \cite{jiang_mistral_2023}, a 7.3 billion parameter model designed to provide a balance between performance and efficiency.
\end{itemize}
\camred{Except for Flan-T5, the models we use in our research were not instruction-tuned during pre-training. This allows us to test the abilities of base LMs to process underspecification.}

\section{Detecting Semantic Underspecification} \label{sec:exp1}
We test whether LMs recognize that a sentence is more or less specified 
by \camred{looking at the perplexity it assigns to a prompt comparing the degree of (under)specification of two embedded sentences}. %explicitly asking them via a prompt. 
This task seeks to replicate the process of assessing human speakers' understanding of this phenomenon by asking them to provide a metalinguistic assessment. \camred{We use a perplexity-based approach, rather than evaluating the generative behavior of the models in response to a prompt, as there is growing evidence that prompt-based approaches are not suitable for this purpose \cite{hu-levy-2023-prompting}. For comparison, we include a preliminary experiment exploring model-generated responses
in~\Cref{sec:MCQ_experiment}}.
% This task aims to simulate how one would test human speakers' knowledge of this phenomenon: by asking them to provide a high-level judgment.

\subsection{Experimental Setup}
%\camred{TODO: make clear we compute perplexity at last token}
We create inputs of the form ``This is an underspecified sentence: \texttt{[sentence1]}. This is its more specified counterpart: \texttt{[sentence2]}'', where \texttt{[sentence1]} and \texttt{[sentence2]} are a pair of sentences from DUST. For each pair, we create a version of the input where the sentences are correctly labeled as under- and more specified and one where their labels are switched.

If the models can recognize underspecification, 
we would expect that the inputs where the specification labels are correct would receive a higher probability / lower perplexity than the same input with incorrect labels. That is, the input

\begin{quote}
    This is an underspecified sentence: `\textcolor{blue}{Andrei left the chair with a green telescope}'. This is its more specified counterpart: `\textcolor{red}{Andrei left the chair on which lay a green telescope}'.
\end{quote}

\noindent should be judged by LMs as more likely than the input where the blue (underspecified) and red (more specified) sentences are switched. \camred{We test this by computing, for a given prompt, the product of the model perplexities assigned to each token in it.}

To ensure prompt diversity, and because models may not have been exposed to terminology such as ``underspecified'' during training, we also use alternate versions of our prompts, where ``under-'' and ``more specified'' are replaced by ``(un)ambiguous', or ``contain (little/a lot of) (information/detail)''. We also create prompts that reverse the order in which the under- and more specified sentences are presented.\footnote{For more example inputs, see \Cref{sec:prompt_examples_exp1}.} In total, we create 33,968 input pairs: 2,123 sentence pairs $\times$ 4 prompt variants $\times$ 4 orders. Then, for each model and input pair, we record whether the model correctly assigns a lower perplexity to the specification-matched input than to its mismatched counterpart.

\begin{figure}[h]
    \centering
    \includegraphics[width=\columnwidth]{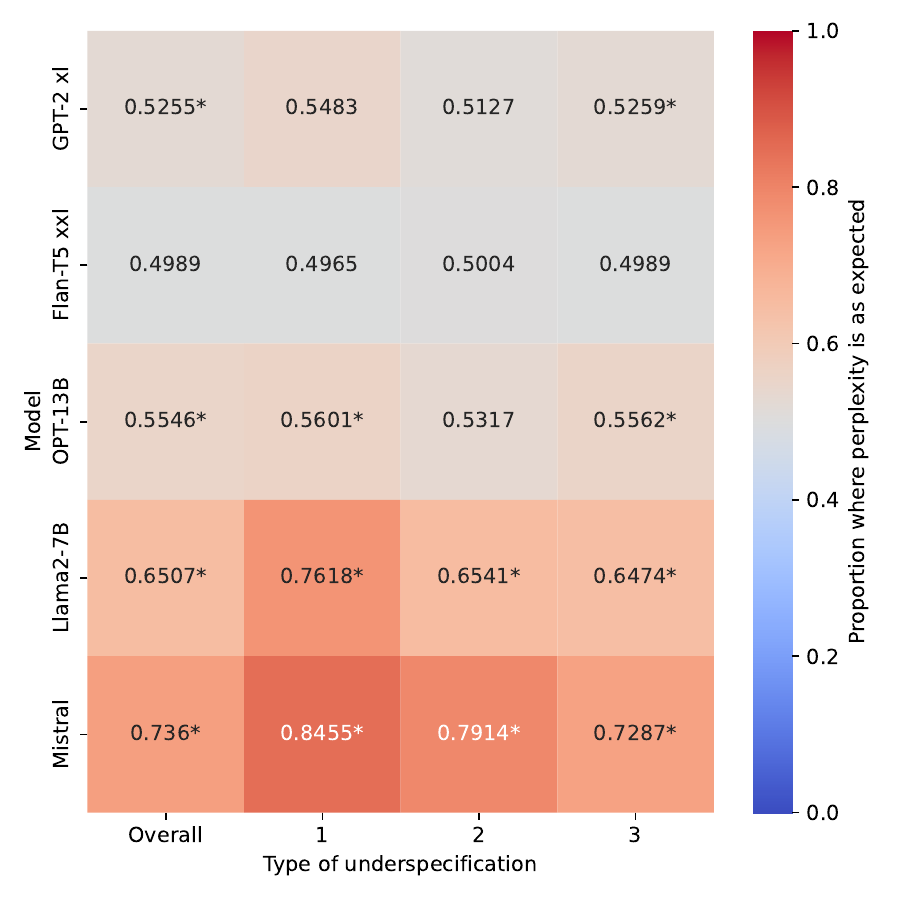}
    \caption{Proportion of specification-matched inputs that receive lower perplexity than their specification-mismatched counterparts; higher is better. %, averaged over phrasings and orders of prompts. 
    Asterisks (*) indicate performance significantly ($p < .05$)
    different from
    % above or below 
    chance (50\%). The leftmost column, \textit{Overall}, reports the proportion computed over the whole dataset.}
    \label{fig:exp1_heatmap}
\end{figure}

%-- Larger models are able to do (significantly) better
%-- (Significant) difference between domains of \citeauthor{egg_semantic_2010} for models that are able to do well.
%-- Effect smaller than sentiment sanity check

\paragraph{Sanity check}
Our experimental setup aims to determine if models can recognize underspecification when prompted to do so; however, prior work suggests that models may struggle to understand their prompts \cite{webson-pavlick-2022-prompt}. So, we first verify the soundness of our setup by using it to test models in an easier domain: sentiment.
%To verify the soundness of our experiment, we first perform it in a different domain: sentiment. 
We gather ``very positive'' and ``very negative'' sentences from the SST-5 dataset \cite{socher-etal-2013-recursive}, and insert them into prompts of the form ``This is a positive sentence: \texttt{[sentence1]}. This is a negative sentence: \texttt{[sentence2]}''. All models assign lower perplexity to sentiment-matched inputs 
with at least 65\% and an average of 75\% accuracy
%with at least 94\% accuracy 
(see \Cref{sec:sentiment_perplexity} for details). This indicates that current LMs can be tested using this experimental setup. 

\begin{table*}[t!]
    \centering
    \begin{tabular}{l | l | l | l | l}
         \textbf{Phenomenon} & \textbf{Sentence} & \textbf{OPT} & \textbf{Llama} & \textbf{Mistral}  \\
         \hline
         Logical Form & Danny approached Andrei; Yevgeni, too & 0.25 & 0.75 & 1\\
         Ellipsis & Andrei and Danny put down a yellow chair & 0.5 & 0.75 & 1\\
         \hline
         PP attach. amb. & Andrei approached the person with a yellow bag & 0.5 & 0.75 & 1\\
         VP attach. amb. & Danny looked at Andrei moving a green chair & 0.5 & 0.75 & 1\\
         Conjunction amb. & Andrei and Danny held the yellow bag and chair & 0.75 & 0.75 & 1 \\
         \hline
         Referential amb. & Although they ran at about the same speed, & 0.25 & 0.25 & 0.25\\
         &  Sue beat Sally because she had such a bad start & & & \\
         \textit{Added compound} & Get yourself a flannel shirt and wear it & 0.75 & 0.75 & 0.75 \\
         & over a plain tee shirt. & & & \\
         \textit{Fused head} & This means you have broken the seal and & 0.5 & 1 & 0.75\\
         & can now twist off the lid. & & & \\
         \textit{Implicit reference} & 4. Do not slurp. & 0.75 & 0.75 & 1 \\
         \textit{Metonymic ref.} & Think about your plant's activity & 0.25 & 1 & 1\\
    \end{tabular}
    \caption{Underspecified sentences and proportion of examples (across prompts and orderings) where the specification-matched input received lower perplexity than its mismatched counterpart for OPT-13B, Llama2-7B, and Mistral 7B; higher is better. Horizontal lines demarcate the types of underspecification as per \citeposs{egg_semantic_2010} taxonomy. Phenomena containing underspecification without ambiguity are \textit{italicized}. `amb.' stands for ambiguity, `ref.' for reference.
    }
    \label{tab:qualitative_1}
\end{table*}

%\begin{table*}[t!]
%    \centering
%    \begin{tabular}{l | l | l | l | l}
%         \textbf{Type} & \textbf{Sentence} & \textbf{OPT-13B} & \textbf{Llama2-7B} & \textbf{Mistral 7B}  \\
%         \hline
%            1 & Danny left Yevgeni; also Andrei & 0.625 & 0.8125 & 0.875\\
%            & Yevgeni approached Andrei; also Danny & 0.6875 & 0.6875 & 0.8125\\ \hline
%            2 & Andrei looked at Danny moving a blue telescope & 0.5 & 0.625 & 0.8126\\
%            & Andrei left the person with a green chair & 0.5 & 0.625 & 0.6875\\ \hline
%            3 & Pay the fee & 0.6875 & 0.5 & 0.875\\
%            & Do this with force, and the can should be smashed  & 0.4375 & 0.5 & 0.5\\
%    \end{tabular}
%    \caption{Underspecified sentences and proportion of examples (across prompts and orderings) where the specification-matched input received lower perplexity than its mismatched counterpart; higher is better.
    % Highest numbers in \textbf{bold}.
%    }
%    \label{tab:qualitative_1}
%\end{table*}

\subsection{Results}
\paragraph{Newer models perform better}
%\camred{TODO: rephrase this section to better make our point: that models trained on more data may eventually be able to acquire better underspecification processing abilities}
Our results (\Cref{fig:exp1_heatmap}) indicate that some LMs can recognize underspecification. All models besides Flan-T5 do so at a rate significantly higher than chance; Flan-T5's
poor performance may be due to its architecture, i.e., a fine-tuned encoder-decoder model and not a decoder-only model trained on causal language modeling like the others.
% performance may be because it is a fine-tuned encoder-decoder model, not a decoder-only model trained only on causal language modeling. 
Stronger models more often prefer specification-matched prompts:
Mistral, 
%the best-performing model across the board, 
\camred{which performs significantly better than all other models across the board ($p < .001$),}
achieves an overall accuracy of 0.74.
% , with a peak of 0.85 in type 1. 
The second-best model, Llama 2, lags behind Mistral by almost 10 accuracy points, with an overall accuracy of 0.65. In turn, this LM \camred{significantly ($p < 0.001$) }outperforms OPT (0.55) and GPT-2 (0.53) by as many accuracy points, indicating a clear ranking between the various models.

% OPT outperforms GPT-2, and the similarly strong Llama 2 and Mistral achieve roughly the same performance. 

% In absolute terms, the models generally achieve a (much) lower accuracy on this task compared to the control experiment with sentiment. We argue that this is likely due to models' difficulties in recognizing and identifying underspecification, rather than prompt-related challenges.  

Given the models' high performance in the control experiment with sentiment, 
% we argue that 
the generally lower accuracy observed here is likely due to models' difficulties in recognizing and identifying underspecification, rather than prompt-related challenges.

% Our results (\Cref{fig:exp1_heatmap}) indicate that stronger models more often prefer specification-matched prompts. However, we can also see that there are differences in performance in this task between the different domains of underspecification. In fact, the domain at which models perform best differs per model. This suggests that although better-performing models are usually better at recognizing underspecification when explicitly prompted, this is not necessarily true for all facets of underspecification.

\paragraph{Results vary across types} Even for top models, performance across different types of underspecification is not uniform, with gaps of up to 12 accuracy points between the best- and worst-performing types.
Mistral, for example, achieves a peak in performance on type 1 (0.85), followed by type 2 (0.79), and type 3
% the most challenging one 
(0.73). For the second best-performing model, Llama 2, type 1 is also the easiest (0.76). However, in contrast with Mistral, types 2 and 3 are equally challenging; Llama 2 achieves similar performance (0.65) on both. These different patterns suggest that the models not only differ in their quantitative ability to perform the task but also in the types of errors they make.

% Moreover, while OPT, Llama 2, and Mistral perform best on the first type of underspecification, the second-best-performing type differs by model.

\paragraph{Qualitative analysis}
%\camred{TODO: expand qualitative analysis}
To shed light on the cases where each model succeeds and fails, we
% the results achieved by the models, we 
conduct a qualitative analysis on a handful of samples, i.e., \camred{one underspecified sentence per linguistic phenomenon, }%two underspecified sentences per DUST type, 
with the best-performing Mistral, Llama 2, and OPT LMs.
% of the results. %We focus on types 1-3, excluding type 4 due to models' near-random performance on it.\footnote{We hypothesize that these results occur because that type does not consist of minimal pairs; though one item of the pair does include a homonymic expression, this does not guarantee that it is overall less specified than its counterpart.} 
The actual sentences considered in this analysis can be found in \Cref{tab:qualitative_1}.
%\camred{We find that although } Mistral is consistently the best-performing model in these examples \camred{the difference between it and the other models, as well as the difference amongst those other models, differs per linguistic phenomenon.}

\camred{Among types 1 and 2 of underspecification, we find that for almost all phenomena, the qualitative inspection closely mirrors the quantitative results reported in ~\Cref{fig:exp1_heatmap} -- Mistral is consistently better than all other models, and Llama 2 outperforms OPT. However, this is much less the case for the \textit{conjunction scopal ambiguity}, where both OPT-13B and Llama2-7B perform at a similar level and are much closer to Mistral. This is in line with the overall quantitative trends in~\Cref{sec:exp1_phenomena}. We hypothesize that this may be because the underspecified sentences displaying this phenomenon can be considered difficult to parse. If true, this would suggest that LMs may use some notion of sentence complexity (e.g. the difficulty to parse it) as a stand-in for underspecification.}

\camred{We also observe that, for referential ambiguity, performance for all models is very low. This may be because most sentences containing referential ambiguity are part of the Winograd Schema Challenge dataset, which may be part of the training data of the tested LMs. As an effect of this, the perplexity assigned to these underspecified sentences may be lower than that of the more uncommon control sentences.}
%As reported in \Cref{tab:qualitative_1}, Mistral is consistently the best-performing model in these examples---it achieves the highest proportion of correctly assigned perplexities. Moreover, the numbers closely align with the patterns reported in~\Cref{fig:exp1_heatmap}: for both Mistral and Llama 2, type 1 is the easiest, followed by types 2 and 3.
% At the same time, and in line with the quantitative results reported in~\Cref{fig:exp1_heatmap}, the model achieves overall higher numbers for type 1 compared to type 2, and in turn for type 2 over type 3. While the same is true for Llama 2, OPT exhibits a different pattern; in these few examples, indeed, type 1 is the easiest, followed by type 3 and type 2. 
%Taking a closer look at the examples, it is worth noting that, (1) for ``Pay the fee'', Llama 2 does not recognize it as more underspecified than its counterpart ``Pay the licensing fee'' (not reported in the table), while OPT and even more Mistral do; (2) for ``Do this with force, and the can should be smashed'', all models similarly struggle; none of them recognize the sentence ``Do this part with force, and the can should be smashed'' (not reported) as its more specified counterpart.

% The results of our analysis suggest that the models that behave as expected more often do not do so for one specific type of sentence in particular. Rather, these models perform better for a large number of sentences across the entire type of underspecification.

\paragraph{Takeaways and discussion} Overall, our results suggest that modern LMs can moderately identify underspecification if explicitly asked to do so via prompting. \camred{In particular, the observation that newer models perform better suggests that pre-training with more and better data, using more parameters, and relying on various (even if minor) architectural improvements could eventually lead to models that can accurately recognize underspecified language.}

However, our current approach does involve significant prompting, which has several disadvantages when evaluating LMs' linguistic abilities. Model performance is highly sensitive to the specific prompt used~\cite{mizrahi2024state}. Moreover, previous work has shown that using meta-linguistic prompts, which explicitly ask for a model's linguistic judgment, often yields different results than evaluating linguistic tasks using naturalistic data~\cite{hu-levy-2023-prompting}. While our experiments do not directly ask models if a given sentence is underspecified (we instead compare two versions of the same sentence), our inputs still do not reflect naturalistic data. Our second experiment is motivated by the need to account for this issue. 

% In our second experiment, we design a second experiment to work around this concern.

%It should be noted that these results do not fully prove that LMs truly understand the concept of underspecification. For example, \citet{webson-pavlick-2022-prompt} argue that models do not (fully) understand the meaning of their prompts. Instead, any prompt would be able to increase a model's performance on such tasks. Although the difference in accuracy between the sentiment task and the underspecification task suggests that this is not the \emph{only} cause for the models' performance, it could still play a role. For this reason, we conduct a second experiment that makes less use of prompts and leaves implicit the fact that the sentences are underspecified.

\section{Interpreting Underspecified and Specified Sentences} \label{sec:exp2}

The results of our first experiment suggest that some LMs may be able to recognize underspecification when asked about it.
However, as discussed above, there is no guarantee that the results obtained using metalinguistic prompts are indicative of the actual capabilities of LMs. 
% in dealing with linguistic phenomena present in naturalistic data. 
In the second experiment, we
use a more naturalistic setting where underspecification is not mentioned in the prompt.
% However, to correctly deal with underspecification, models should also be able to reason with underspecification even when the underspecification is not explicitly mentioned. Our second experiment investigates this ability.

\subsection{Experimental Setup}
% To test LMs' implicit underspecification processing abilities, w
We create two specified versions of each DUST sentence originating from the LAVA dataset, corresponding to the possible readings of the original sentence. We also create two continuations to the sentence, which again correspond to distinct readings of the original. Each continuation is compatible with the underspecified sentence, but only one of the fully specified sentences. We slightly adjust the LAVA sentences to make them more correct.\footnote{Uses of `put-down' and `picked-up' were changed to `put down' and `picked up', and sentences of the form ``\texttt{NNP}$_{1}$ \texttt{V} \texttt{NNP}$_{2}$. Also \texttt{NNP}$_{3}$.'' were changed to ``\texttt{NNP}$_{1}$ \texttt{V} \texttt{NNP}$_{2}$; \texttt{NNP}$_{3}$ too.''} 
% For an example, see \Cref{tab:continuations_example}.

\begin{comment}
\begin{table}[t!]
    \centering
    \begin{tabular}{l | l}
         \textbf{Underspecified sentence} & Andrei left Yevgeni;\\
         & Danny too\\
         \hline
         \textbf{Specified Counterpart} & Andrei left Yevgeni;\\
         & Danny also did\\
         \hline
         \textbf{Correct Continuation} & Danny left Yevgeni\\
         \hline
         \textbf{Incorrect Continuation} & Andrei left Yevgeni \\
    \end{tabular}
    \caption{An underspecified sentence, and correct and incorrect continuations for its more specified counterpart.}
    \label{tab:continuations_example}
\end{table}
\end{comment}

\begin{figure}[t!]
    \centering
    \includegraphics[width=\columnwidth]{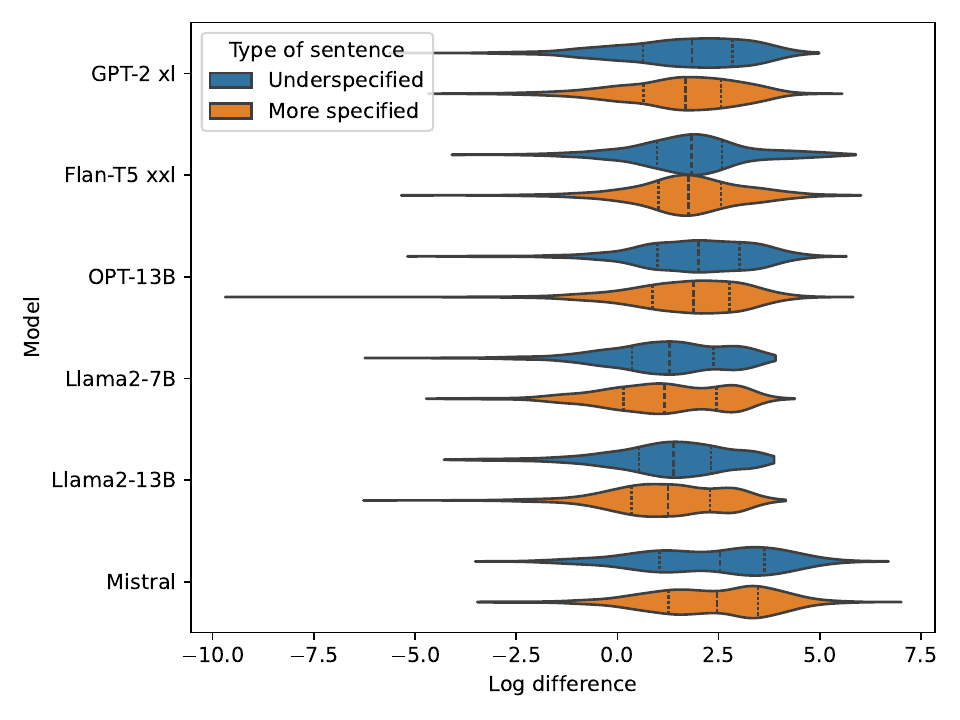}
    \caption{Log difference of perplexity of correct and incorrect continuations for underspecified (blue) and more specified (orange) sentences, averaged over prompts. The underspecified condition is not significantly lower than the more specified condition for any model. 
    }
    \label{fig:exp2_violinplot}
\end{figure}

We then embed these sentences and continuations in simple templates like ``\texttt{[sentence]}. That is, \texttt{[continuation]}''. Given an underspecified sentence, where both continuations are equally plausible, LMs should ideally assign roughly equal perplexity to both.\footnote{Underspecified sentences could still have a more frequent ``default reading'', making one continuation more likely in human speakers' interpretation. \camred{For example, when resolving referential ambiguity, it has been documented that a reader might default to the first possible referent encountered in the text. Alternatively, it has been proposed that, when parsing a sentence, speakers could leave ambiguities unresolved if resolving is not strictly necessary \cite{swets_underspecification_2008}. Since we consider both sentence readings in our experiment, we hypothesize that the effects of such default readings cancel out and therefore do not affect our results.}} 
%\camred{TODO: expand on this discussion; find paper to cite}
For more specified sentences, in contrast, only one interpretation is possible; the corresponding continuation should therefore receive a higher probability than the incompatible one.

For example, given the underspecified sentence ``Danny looked at Andrei with a telescope,'' the continuations ``That is, Andrei had a telescope,'' and ``That is, Danny had a telescope,'' should be similarly likely. However, given the sentence ``Danny looked at Andrei, who had a telescope,'' the first continuation would be more likely. 

We experiment with connecting the sentence and continuation in different ways; besides inserting ``That is,'' between them, we also try using no connector (``\texttt{[sentence]}. \texttt{[continuation]}''), and stating a more explicit connection: ``\texttt{[sentence]}. Therefore, it is more likely that \texttt{[continuation1]} than \texttt{[continuation2]}.''.

We record the absolute value of the difference in the perplexities assigned to each continuation given an underspecified sentence. We expect models will generally have only weak preferences between continuations when given underspecified sentences, leading to smaller absolute differences; specified sentences should have larger differences. For each specified sentence, we also record if LMs prefer the plausible continuation over the implausible. 

\subsection{Results}
%\camred{Include significance test between models}
\begin{figure}[t!]
    \centering
\includegraphics[width=\columnwidth]{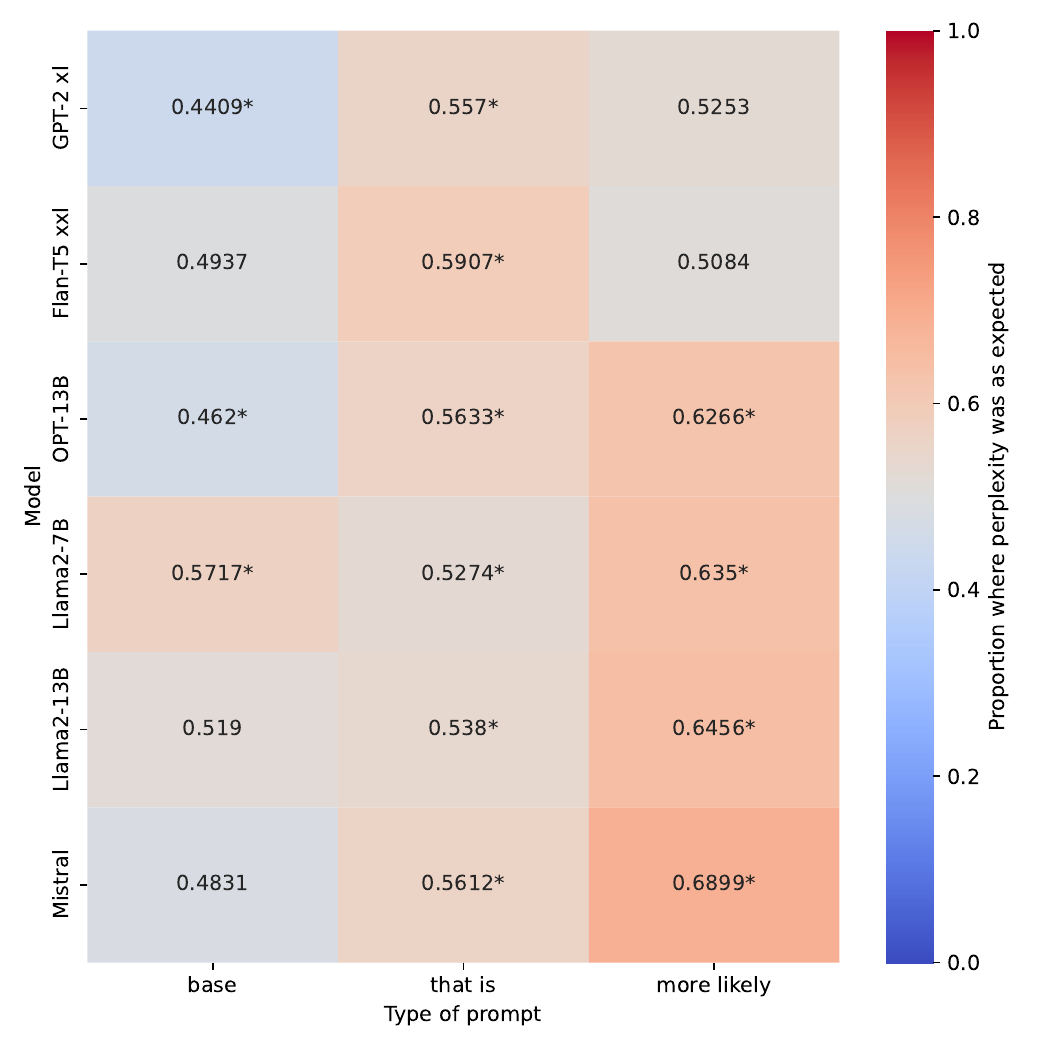}
\caption{Proportion of inputs where the plausible continuation to a specified sentence received lower perplexity; higher is better. Asterisks (*) indicate performance that differs significantly ($p < .05$) from chance (50\%).
    }
    \label{fig:exp2_heatmap}
\end{figure}

\paragraph{LMs do not have stronger preferences toward specified, rather than underspecified sentences}
Our results (\Cref{fig:exp2_violinplot}) indicate that the tested models do not interpret underspecified sentences and their specified counterparts correctly: there is no statistically significant difference between the differences in perplexities assigned to continuations of underspecified sentences, and differences in those assigned to continuations of more specified sentences. This suggests that models might incorrectly assign one single interpretation to underspecified sentences, or they might also fail to assign only one interpretation to more specified sentences; a combination of these is also possible.

\begin{table*}[t!]
    \centering
    % \small
    \begin{tabular}{l | l | l | l}
         \textbf{Sentences and Continuations} & \textbf{OPT-13B} & \textbf{Llama2-7B} & \textbf{Mistral}  \\
         \hline
            Andrei looked at Danny holding a yellow bag & 20.49& 2.26 & 48.51\\
            Andrei looked at Danny while holding a yellow bag & 13.77& \textbf{2.74} & 32.61\\ 
            $\rightarrow$[Andrei / Danny] had a yellow bag &&&\\
            \hline
            Andrei looked at Danny holding a yellow bag & 5.07 & 2.42 & 12.81\\
            Andrei looked at Danny while holding a yellow bag & \textbf{6.03} & \textbf{5.54} & \textbf{13.32}\\ 
            $\rightarrow$\textit{That is,} [Andrei / Danny] had a yellow bag &&&\\ \hline
            Andrei looked at Danny holding a yellow bag & 1.47 & 0.68 & 2.91\\
            Andrei looked at Danny while holding a yellow bag & 1.34 & 0.23 & \textbf{5.84}\\
            $\rightarrow$\textit{Therefore, it is more likely that} [Andrei...] \textit{than} [Danny...]
    \end{tabular}
    \caption{For each prompt type (base, \emph{that is}, and \emph{more likely}), an underspecified sentence and one more specified counterpart, along with the absolute difference in perplexities assigned to the continuations of each. The LM is right when this number 
    % (difference in perplexity between continuations) 
    is higher for the second, more specified, than the first, underspecified, sentence (\textbf{bold} cases).}
    \label{tab:qualitative_2}
\end{table*}

% In other words, the models are not able to determine that the incorrect continuation only follows from the underspecified sentence, or in other words, do not notice that the underspecified sentences are underspecified.

% \paragraph{Do models exhibit the expected preferences given specified sentences?}
\paragraph{The more explicit the prompt, the better the results}
Our results (\Cref{fig:exp2_heatmap}) indicate that the type of prompt used heavily influenced the degree to which models preferred the correct, plausible continuation; however, unlike in experiment 1, performance is poor overall. In the base case, where no prompts were used,
only Llama 2 7B performs significantly above chance. With the ``that is'' prompt, model performance improves, though there is no clear trend in which models perform best. It is only with the most extensive prompt that we recover both better performance and the trend from experiment 1, where newer models perform better, with Mistral once again performing significantly ($p = .002$) better than all other models. 

This trend, where more explicit prompts yield better results, is surprising: prior work \cite{hu-levy-2023-prompting} suggested that metalinguistic prompts might more poorly capture LM capabilities, underestimating them. We hypothesize that this could occur because the continuations we crafted are sometimes more like conclusions that follow from the first sentence (as in NLI), and less like genuinely probable continuations. We note, however, that other work has observed that both LMs and humans sometimes default to NLI when given sentence pairs without instructions \cite{webson2023are}.

\paragraph{Qualitative analysis}
%\camred{TODO: expand qualitative analysis}
We also examine the degree to which models assign less univocal interpretations to underspecified sentences than their specified counterparts. \Cref{tab:qualitative_2} shows the absolute difference in perplexity assigned to the continuations of one under- and one more specified sentence, by prompt and model. On this particular example, the models unanimously assign the more specified sentence a more univocal reading only on the ``that is'' prompt; on the others, they disagree, echoing the noisiness of our quantitative results.

%This noisiness is also reflected in the differences between linguistic phenomena and between type of prompt. We find that for referential ambiguity and VP attachment ambiguity, the proportion of inputs where the plausible continuation to a specified sentence received lower perplexity is considerably higher. However, due to the small amount of sentences containing these linguistic phenomena in our dataset when considered separately, we are apprehensive in drawing conclusions from this observation.

This noisiness is also reflected in the differences between linguistic phenomena and between type of prompt. We find that for referential and VP attachment ambiguity, a greater proportion of inputs resulted in the correct continuation receiving lower perplexity. This seems to suggest that correctly interpreting sentences containing these two phenomena is handled better by the models, although this differs greatly per model and type of prompt. However, we note that relatively few sentences contain these linguistic phenomena, limiting our ability to draw strong conclusions from this observation.

\section{Discussion}
Underspecification, much like ambiguity~\cite{liu_were_2023}, remains a challenging phenomenon for LMs. Older LMs, such as GPT-2, perform near chance level at recognizing underspecification; newer models, such as Llama 2 and Mistral, perform much better, but still leave ample room for improvement. Processing sentences containing underspecification is an even harder task for LMs. Models seem to fail to
% Models do not 
% seem to 
recognize when underspecified sentences license continuations that their more specified counterparts do not; moreover, their interpretations of more specified sentences are often incorrect.

The striking difference between the results of our two experiments highlights the importance of carefully choosing a setup when evaluating model capabilities. In the first, metalinguistic prompts elicited good underspecification judgments from high-performing models. But, in surprising contrast to previous work \cite{hu-levy-2023-prompting}, testing models' ability to process underspecification in a more naturalistic setting led to lower performance. This is important: LM use cases involving underspecification will most likely involve processing underspecification, rather than identifying it upon explicit request. Our second experiment may thus be a better indicator of LMs' practical abilities. 
However, future work may be needed to compare LM processing of underspecified sentences to results of human studies, which have shown that speakers do have default interpretations of such sentences \cite{Kurtzman1993ResolutionOQ,Dwivedi2013InterpretingQS}.

By introducing DUST and studying underspecification in LMs as distinct from ambiguity, we 
have taken
the first step towards evaluating LMs' performance on a commonplace but understudied phenomenon that can affect LM behavior. Our findings show
% In conclusion, our research 
that current LMs are limited in their ability to deal with underspecification, especially in genuine communicative scenarios. Hence, a thorough evaluation of the abilities of LMs should include (various facets of) underspecification, unlike current benchmarks, in which ambiguous and underspecified sentences are often systematically excluded. We hope that our research further showcases the relevance of underspecification as a direction of research in the study of language models. %To this end, we will release our code and data upon acceptance.

\section*{Limitations}

%\camred{TODO: mention machine translation as a testbed for future work (perhaps as an example of research on a more realistic task?)}

DUST is arguably a small dataset, and would benefit from expansion. While we considered existing resources and extracted linguistic data with the desired features from those, future work could expand it by collecting new data via human annotation, generation, or other data-driven approaches. This holds particularly true for type 1, which contains much fewer examples than the other types.

While the present work performs an in-depth evaluation of how LMs behave when faced with semantic underspecification, our research does not explore the inner mechanisms that underlie this capability. We acknowledge that doing so would provide complementary evidence that may be needed to shed full light on the phenomenon. \camred{Moreover, research could focus on how LMs handle underspecification in more naturalistic scenarios, e.g., in the context of real-world NLP applications, which is something the current work does not explore.} 
% also attempt to shed light on how LMs handle underspecification when part of an application, which is not something investigated in the current work.}

Our research builds on the formal categorization of semantic underspecification by~\citet{egg_semantic_2010}. While this theoretical framework is both comprehensive and generally suitable for our purposes, we are aware that other theoretical accounts may define the semantic underspecification slightly differently, by including more/less phenomena or by categorizing them according to different criteria. Future work could explore whether and how our findings generalize to other formalizations.

\section*{Ethics Statement}
While this work presents no serious ethical concerns, a general consideration needs to be made about the use of pre-trained LMs. As is commonly acknowledged, these models should be used with caution as they could perpetuate harmful biases present in their training data. Furthermore, there is a risk that they will generate false or misleading output. In our work, we minimize these risks as we do not use the LMs to generate output, but only to score the plausibility of sentences fed as input. At the same time, we are also aware that some biases may also be present in the linguistic data we used.

\camred{
\section*{Acknowledgements}
%TODO: acknowledgements. 
Some of the contents of this work are based on FW's Master's thesis. Full acknowledgments from the first author can be found therein. We thank the anonymous ARR reviewers for their valuable feedback and the members of the Dialogue Modelling Group at the University of Amsterdam for their insightful comments. MH's research is supported by an OpenAI Superalignment Fellowship.}

\bibliography{bib_clean}

\appendix

\section{Type 4 of underspecification}
\label{sec:type4}
%\camred{TODO: expand on this discussion}
In an attempt to collect sentences containing this type of underspecification, we collect sentences containing homonymic expressions from a sample of English Wikipedia \citep{wikisent}. We do so by selecting sentences that contain any homonym from a list of 100 homonyms, selected by~\citet{maciejewski_relative_2016} based on linguistic principles, dictionary entries, and subjective ratings. We created more specified counterparts by selecting random sentences from the that contain none of the homonyms from the list. We collect a total of 980 sentence pairs; note that unlike other pairs in DUST, these are not minimal pairs. One (partial) sentence of this type (though not from our dataset) is
\begin{quote}
    The elderly fish.
\end{quote}
which is underspecified because `fish' can be both a noun or a verb in this context. A more specified counterpart of this (partial) sentence could be
\begin{quote}
    The elderly people fish.
\end{quote}
Note, however, that the reading where `fish' is a noun is not a full sentence, and would only be grammatical when placed in a context (e.g. an enumeration) where it is suitable.

\begin{figure}
    \centering
    \includegraphics[width=\columnwidth]{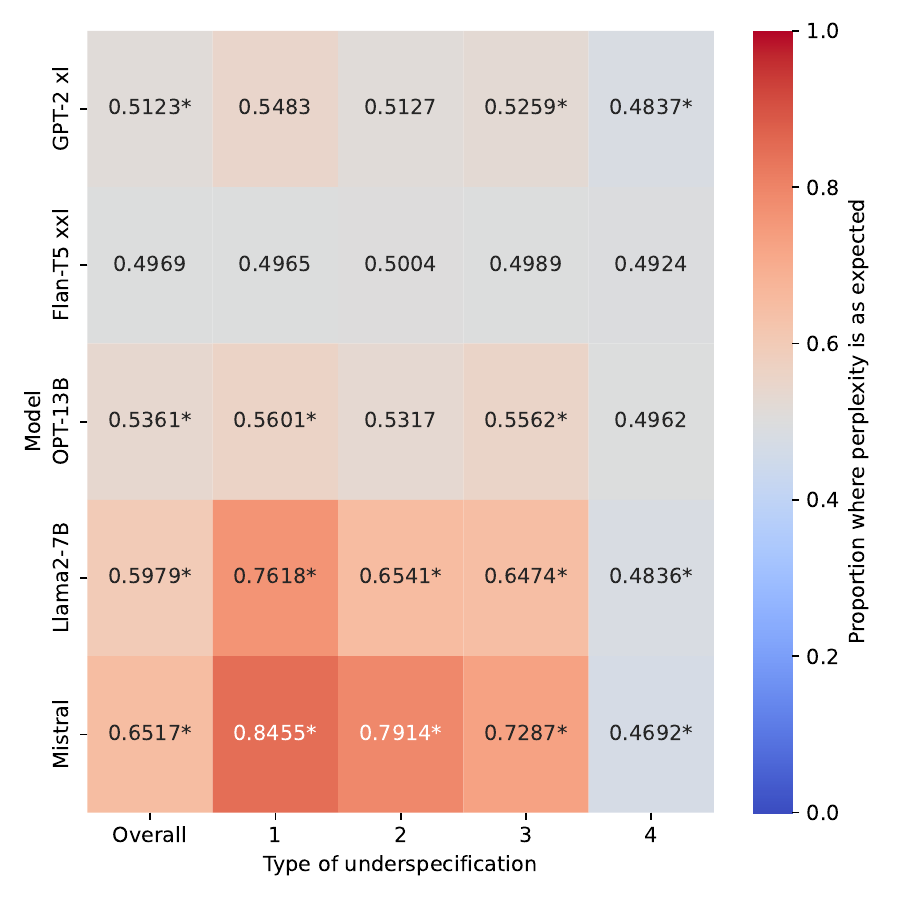}
    \caption{Proportion of specification-matched inputs with lower perplexity than their specification-mismatched counterparts; higher is better. %, averaged over phrasings and orders of prompts. 
    Asterisks (*) indicate performance significantly ($p < .05$) above or below chance (50\%). }
    \label{fig:exp1_heatmap_incl_4}
\end{figure}

Running experiment 1 with these sentences included, we get the results shown in \Cref{fig:exp1_heatmap_incl_4}. We can see that models achieve poor performance on type 4 sentences across models. We hypothesize that this is because that type of underspecification does not consist of minimal pairs; though one item of the pair does include a homonymic expression, this does not guarantee that it is overall less specified than its counterpart. Due to these caveats, we have excluded this type from our dataset.

\section{Dataset Information}
\label{sec:dataset_info}
In this section, we review important licensing and privacy information regarding the datasets composing DUST, as well as DUST itself. DUST is composed of data from 3 datasets.
\begin{itemize}
    \item LAVA \citep{berzak_you_2015}, available at \url{https://web.mit.edu/lavacorpus/}, was released with an unclear (potentially open-source) license. LAVA contains images of the authors, which DUST does not include. However, the names used in the sentences in LAVA refer to its authors. These sentences contain no other personally identifiable or offensive content.
    \item CLAIRE \citep{roth_semeval-2022_2022}, available at \url{https://github.com/acidAnn/claire}, is composed of WikiHow articles released under a CC BY-NC-SA 3.0 license, and was itself released under the same license. We did not filter the articles and revisions of which CLAIRE is comprised for personally identifiable or offensive content. 
    \item The Winograd Schema Challenge dataset \citep{levesque_winograd_2012}, available at \url{https://huggingface.co/datasets/winograd_wsc}, was released under a CC BY 4.0 license. All sentences were created by experts, and do not contain personally identifiable or offensive content.
\end{itemize}
DUST is a non-commercial dataset which, like all of its component datasets, is intended for research purposes. We have also provided attribution to the creators of the component datasets, allowing it to be released in accordance with all of the licensing terms of these component datasets. Owing to the WikiHow data contained within, we also release DUST with a CC BY-NC-SA license.

\section{Model and Experimental Details}
\label{sec:model_exp_details}
In this section, we provide further information regarding our models and experiments. We use the HuggingFace transformers implementations of all models, available at \url{https://huggingface.co/models}. We also use the HuggingFace weights for all models except Llama 2, which must be downloaded separately at \url{https://llama.meta.com/llama-downloads/}. To compute perplexities, we use LM-PPL: \url{https://github.com/asahi417/lmppl}.

We ran these experiments on compute nodes equipped with Nvidia A100 GPUs (40GB RAM); for all models but Llama 2 13B, one such GPU should suffice. The runtime of our experiments is no more than 5 GPU days. 

\section{MCQ experiment}
\label{sec:MCQ_experiment}
%\camred{TODO: include MCQ experiment}
\camred{In this experiment, we test whether language models can recognize semantic underspecification by explicitly asking them to generate an answer about the underspecification of a sentence pair. In practice, we prompt GPT2-xl, OPT-13B, and Llama2-7B to generate a response using the following prompt:}

\begin{quote}
    Here are two sentences. A: `\textcolor{blue}{Andrei left the chair with a green telescope}'. B: '\textcolor{red}{Andrei left the chair on which lay a telescope}'. Which one of these is more semantically underspecified? Please respond by outputting only A or B. Answer: 
\end{quote}
\camred{where the red and blue sentences are replaced by either an underspecified or its corresponding
control sentence from the dataset. The order in which the two sentences are placed is randomized to prevent bias in the model from unduly influencing the final accuracy. Each sentence pair in the dataset is included in a prompt once. In ~\Cref{tab:MCQ}, we report model accuracy and number of A, B, or other responses.}

\begin{table}[h]
    \centering
    \begin{tabular}{l | c c c c}
        \textbf{model} & \textbf{acc.} & \textbf{\#A} & \textbf{\#B} & \textbf{\#other} \\
        \hline
         \textbf{GPT2-xl} & 0.31
         % 0.3089 
         & 696 & 638 & 789   \\
         \textbf{OPT-13B} & 0.49
         % 0.4899 
         & 2105 & 18 & 0\\
         \textbf{Llama2-7B} & 0.48
         % 0.4790 
         & 888 & 1138 & 97\\
    \end{tabular}
    \caption{MCQ task. Model accuracy and number of responses per class by GPT2, OPT, and Llama 2.}
    \label{tab:MCQ}
\end{table}

\camred{The results show that the models perform very poorly when the task is formulated as a multiple-choice task. While the low accuracy might be a result of the models being unable to do this task---something that could perhaps improve when using instruction-tuned models---the observation that all the models are either biased towards one of the options or unable to consistently answer the prompt with one of the two given options, or both, suggests that they are incapable of performing the task in this experimental setup. This matches earlier findings (e.g., \citealp{hu-levy-2023-prompting}) and validates our decision to perform perplexity-based evaluation over a MCQ-like type of experiment even further.}

\section{Experiment 1 Prompts}
\label{sec:prompt_examples_exp1}

Suppose we have the underspecified sentence ‘Andrei left the chair with a blue telescope’ and the more specified counterpart ‘Andrei left the chair on which lay a blue telescope’. Examples of specification-matched prompts we would then obtain are:
\begin{quote}
    This is an underspecified sentence: `Andrei left the chair with a blue telescope'. This is its more specified counterpart: `Andrei left the chair on which lay the blue telescope'.
\end{quote}
and
\begin{quote}
    This is a sentence that contains a lot of detail: `Andrei left the chair on which lay the blue telescope'. This is a sentence that contains little detail: `Andrei left the chair with a blue telescope'.
\end{quote}
and examples of mismatched prompts are:
\begin{quote}
    This is an ambiguous sentence: `Andrei left the chair on which lay the blue telescope'. This is its unambiguous counterpart: `Andrei left the chair with a blue telescope'.
\end{quote}
and
\begin{quote}
    This is a sentence that contains a lot of information: `Andrei left the chair with a blue telescope'. This is its counterpart that contains little information `Andrei left the chair on which lay the blue telescope'.
\end{quote}
These examples show all variations of phrasing and order of parts

\section{Experiment 1 Results per Phenomenon}
\label{sec:exp1_phenomena}
\camred{In~\Cref{fig:exp1_phenomena}, we report the results of Experiment 1 split by linguistic phenomenon.
\begin{figure*}
    \centering
    \includegraphics[width=\textwidth]{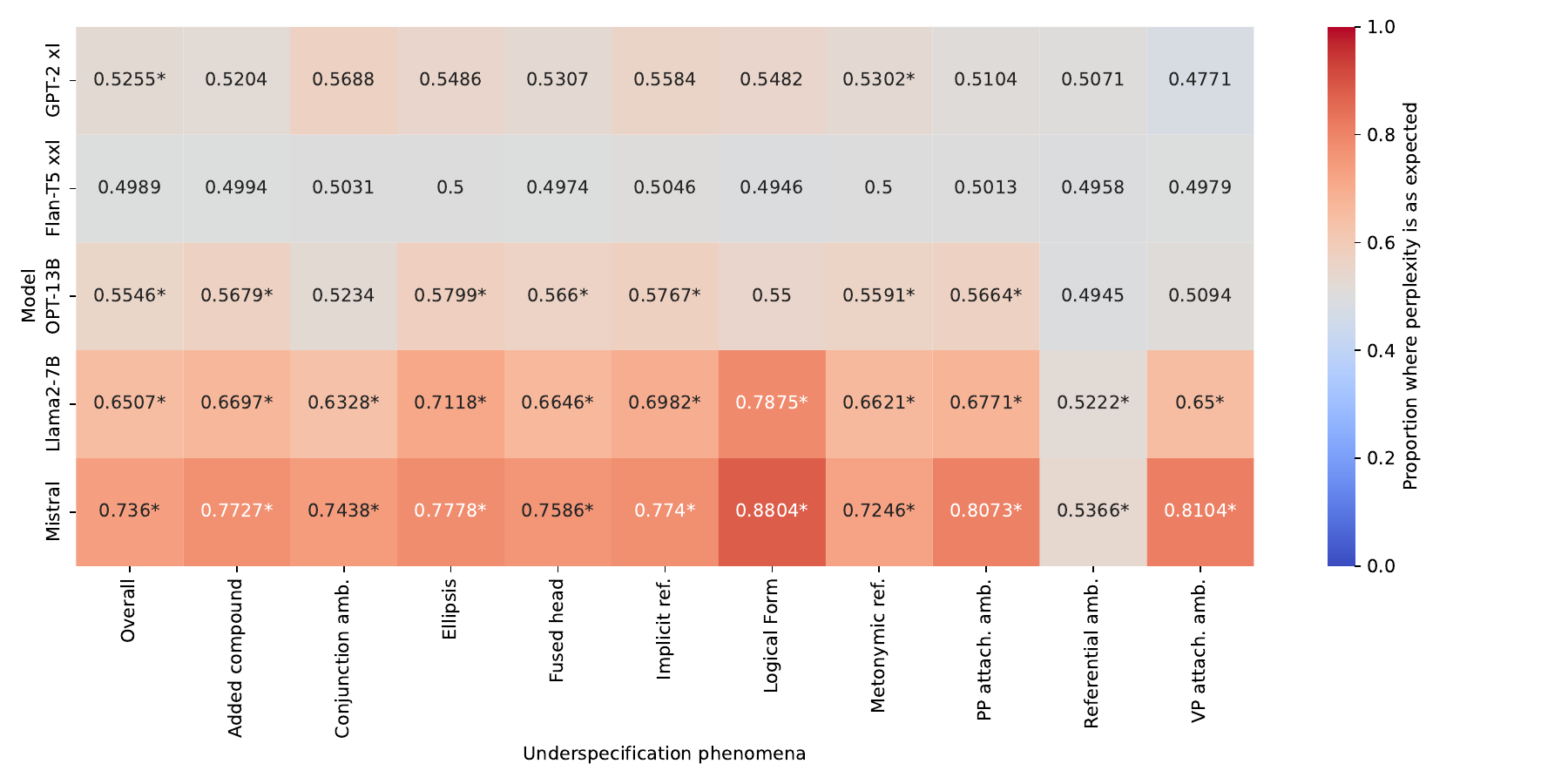}
    \caption{Proportion of specification-matched inputs with a lower perplexity than their specification-mismatched counterparts, split by linguistic phenomenon; higher is better. %, averaged over phrasings and orders of prompts. 
    Asterisks (*) indicate performance significantly ($p < .05$) above or below chance (50\%). }
    \label{fig:exp1_phenomena}
\end{figure*}
}

\section{Sentiment perplexity}
\label{sec:sentiment_perplexity}

\begin{figure}[ht]
    \centering
    \includegraphics[width=\columnwidth]{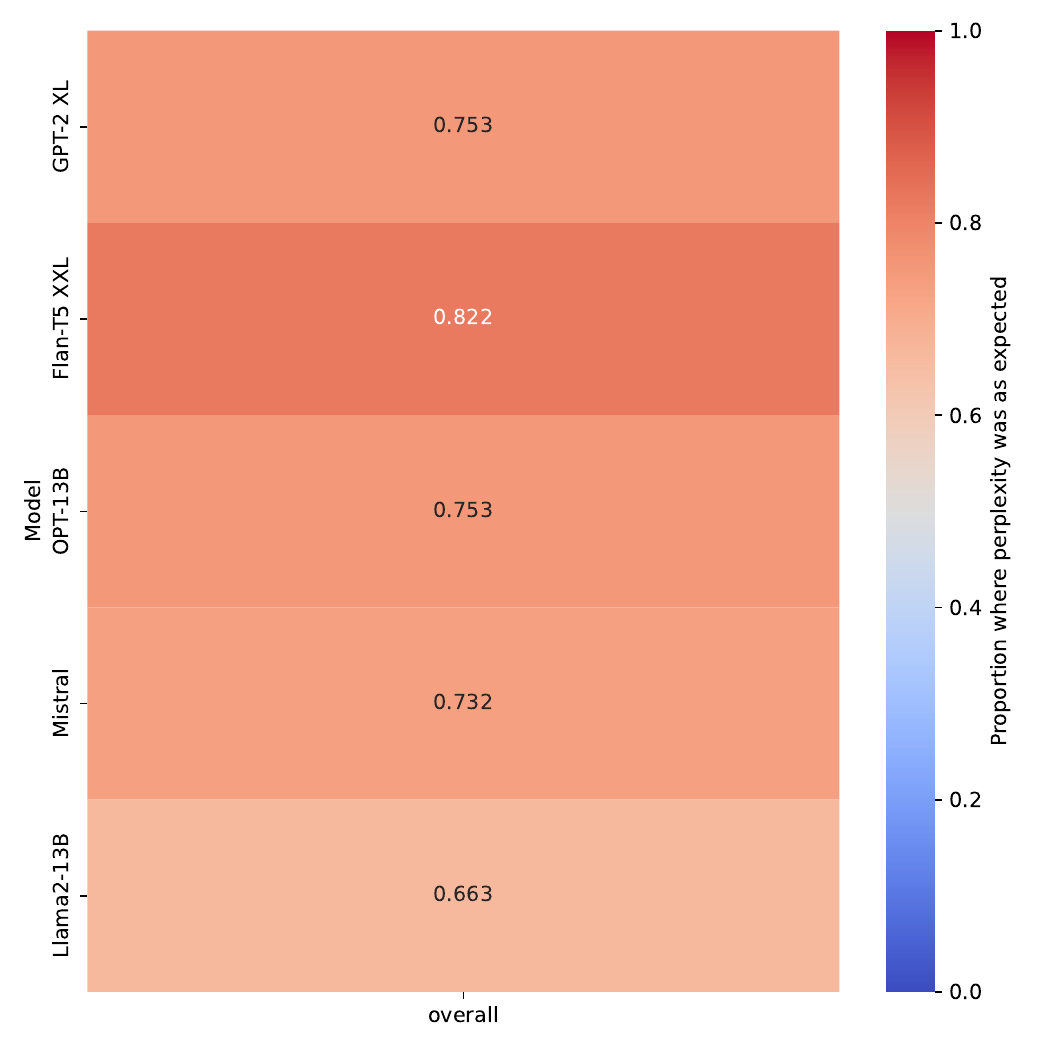}
    \caption{Percentage of sentences from experiment 1 with sentiment instead of underspecification where the perplexity of the correct prompts is lower than that of the incorrect prompts, averaged over orders of prompts}
    \label{fig:sentiment_heatmap}
\end{figure}

To test whether the experimental design of experiment 1 functions correctly, we first ran this experiment with sentiment classification instead of recognition of underspecification as a goal. The models were prompted with prompts of the form ``{prompt1}: `{sentence1}'. {prompt2}:  `{sentence2}'.", where the prompts are of the form "This is a positive/negative sentence" and the sentences are sentences rated `very positive' or `very negative' in the SST-5 dataset \citep{socher-etal-2013-recursive}. The results of this can be seen in \Cref{fig:sentiment_heatmap}.

\section{Experiment 2 Prompts}
\label{sec:prompt_examples_exp2}
Suppose we have the underspecified sentence `Andrei looked at Danny moving a green bag' and the more specified counterpart ‘Andrei looked at Danny who was moving a green bag
’. Examples of specification-matched prompts we would then obtain, from low to high levels of prompting, are:
\begin{quote}
    Andrei looked at Danny who was moving a green bag. Danny was moving a green bag.
\end{quote}
\begin{quote}
    Andrei looked at Danny who was moving a green bag. That is, Danny was moving a green bag.
\end{quote}
\begin{quote}
    Andrei looked at Danny who was moving a green bag. Therefore, it is more likely that Danny was moving a green bag than Andrei was moving a green bag.
\end{quote}
and examples of mismatched prompts are
\begin{quote}
    Andrei looked at Danny who was moving a green bag. Andrei was moving a green bag.
\end{quote}
\begin{quote}
    Andrei looked at Danny who was moving a green bag. That is, Andrei was moving a green bag.
\end{quote}
\begin{quote}
    Andrei looked at Danny who was moving a green bag. Therefore, it is more likely that Andrei was moving a green bag than Danny was moving a green bag.
\end{quote}

\begin{table*}[h]
\centering
    \begin{tabular}{lcccccc}
    Variable & Coefficient & Standard Error & $z$ & $P>|z|$ & $[0.025$ & $0.975]$\\
    \hline
    sen. len. & -0.155 & 0.0623 & -2.3418 & 0.2135 & -0.2775 & -0.0328\\
    avg. AoA. & -0.2616 & 0.2443 & -1.046 & 0.3642 & -0.7405 & 0.2173\\
    \textbf{avg. conc.} & 1.7596 & 0.65 & 2.6663 & \textbf{0.0262} & 0.4855 & 3.034\\
    avg. word freq. & -2.9801 & 2.1483 & -1.394 & 0.1245 & -7.1908 & 1.2307\\
    avg. word len. & 0.0367 & 0.4773 & 0.062 & 0.3468 & -0.8987 & 0.972
    \end{tabular}
    \caption{Average regression coefficients averaged over all tested models with sentence length, average age of acquisition \cite{kuperman_age--acquisition_2012}, average concreteness \cite{brysbaert_concreteness_2014}, average word frequency \cite{norvig2009natural} and average word length as independent variables, and whether the perplexity of the correct continuation of the specified counterpart is higher than that of the incorrect continuation as the dependent variable}
    \label{tab:logregr}
\end{table*}
\section{Could models be detecting surface statistics instead of underspecification?}
To investigate whether the models' ability to interpret underspecification as underspecification correlates with some surface-level statistic of the sentences in the dataset, we fit a logistic regression model with surface-level descriptive statistics about each sentence as independent variables and the model `correctness' from  \Cref{fig:exp2_heatmap} as the dependent variable. The results of this can be seen in \Cref{tab:logregr}.

These results suggest that models are better able to recognize semantic underspecification when the words in the sentence are more concrete. No other surface-level statistic we test shows a significant correlation with the ability of the models to interpret underspecification.

This agrees with intuition: unlike other features like age of acquisition, word frequency or sentence length, concreteness is something humans also associate with (under)specification -- for example, when a speaker wants to make things less underspecified, they might say ``let's make things concrete". However, the fact that models do not correctly interpret underspecified sentences when these sentences are abstract in nature does pose a problem, given the fact that certain types of text (e.g. legal texts or product specification documents) can be very abstract while requiring all potential underspecification to be correctly interpreted.

\end{document}